\documentclass{article}
\usepackage{ijcai23}

\usepackage{times}
\usepackage{soul}
\usepackage{url}
\usepackage[hidelinks]{hyperref}
\usepackage[utf8]{inputenc}
\usepackage[small]{caption}
\usepackage{graphicx}
\usepackage{amsmath}
\usepackage{amsthm}
\usepackage{booktabs}
\usepackage{algorithm}
\usepackage{algorithmic}
\usepackage[switch]{lineno}

\usepackage{multirow}

\title{Truveta Mapper: A Zero-shot Ontology Alignment Framework}

\author{
Mariyam Amir
\and
Murchana Baruah\and
Mahsa Eslamialishah\and
Sina Ehsani$^*$\and
Alireza Bahramali\footnote{This work was done during an internship at Truveta.} \and
Sadra Naddaf-Sh\And
Saman Zarandioon
\affiliations
Truveta, Bellevue, WA 98004, USA
}

\begin{document}

\maketitle
\begin{abstract}
    In this paper, a new perspective is suggested for unsupervised Ontology Matching (OM) or Ontology Alignment (OA) by treating it as a translation task. Ontologies are represented as graphs, and the translation is performed from a node in the source ontology graph to a path in the target ontology graph. The proposed framework, Truveta Mapper (TM), leverages a multi-task sequence-to-sequence transformer model to perform alignment across multiple ontologies in a zero-shot, unified and end-to-end manner. Multi-tasking enables the model to implicitly learn the relationship between different ontologies via transfer-learning without requiring any explicit cross-ontology manually labeled data. This also enables the formulated framework to outperform existing solutions for both runtime latency and alignment quality. The model is pre-trained and fine-tuned only on publicly available text corpus and inner-ontologies data. The proposed solution outperforms state-of-the-art approaches, Edit-Similarity, LogMap, AML, BERTMap, and the OM frameworks presented in Ontology Alignment Evaluation Initiative (OAEI22), offers log-linear complexity, and overall makes the OM task efficient and more straightforward without much post-processing involving mapping extension or mapping repair. 
    We are open sourcing our solution\footnote{\href{https://github.com/sadransh/ontology-matching-framework}{https://github.com/sadransh/ontology-matching-framework}}.
\end{abstract}

\section{Introduction}
Ontology Matching (OM) or Ontology Alignment (OA) is the process of finding correspondence between the entities of two ontologies. The purpose of this process is to unify data from different sources and reduce heterogeneity, making data more viable for research and development \cite{neutel2021towards}. Classical state-of-the-art (SOTA) approaches on OM are based on non-contextual matching, where the model captures lexical similarity but fails to understand textual semantics, which results in ambiguity. On the other hand, with contextual approaches, the objective is to match complex pairs which are lexically different but semantically similar and vice-versa. For example, ``Encephalopathy'' and ``Disorder of brain'' are lexically different but are used in the same context. However, ``Structure of permanent maxillary right second molar tooth'' and ``Structure of permanent mandibular right first molar tooth'' are lexically similar but are semantically different.

Recently, a transformer-based contextual framework using BERT \cite{devlin2018bert}, has been proposed in \cite{he2022bertmap}, which showed promising results compared to other OM systems. {In their approach the existing pre-trained BERT model was fine-tuned to learn the similarity between different terms, and thereby achieve equivalence matching. This process involves computing the similarity of each input term with a large subset of terms in the target ontology, resulting in quadratic complexity. Additionally, the model captures textual context, however, it does not understand the ontology graph structure, which could significantly extend the capabilities of ontologies graph matching.}  

Motivated by the potential of the transformer models for understanding textual semantic context and overcoming the limitations in the existing methods, the present work proposes Truveta Mapper (TM), a novel zero-shot sequence-to-sequence multi-task transformer-based framework for OM, with the capability of learning both the graph-structure and textual semantics of the ontologies. The model is first pre-trained to learn the hierarchical graph structure of ontology and semantics of each class using Masked Language Modeling (MLM), then fine-tuned using class labels and synonyms as input and class hierarchical-ID as the output, capturing the structure of the ontology. As such, we treat OM as a translation task, where the source ontology class is translated to a path in the matching target ontology class in a zero-shot and multitask manner. Proposed approach is based on zero-shot learning and prediction, where ``zero-shot learning'' refers to the ability of the model to make source-to-target predictions without requiring manually labeled cross-ontologies matching pairs, and ``zero-shot prediction''  performs end-to-end mapping from the source to the target without any similarity calculation across the entire/subset target ontology or post-processing like extension/repair. With multi-tasking, a single model is capable of matching different ontologies such as SNOMED to FMA, SNOMED to NCIT, and so on, and takes advantage of transfer learning as well.

In this work, empirical comparison is made with the state-of-the-art lexical matching approaches and the recent contextual models presented in \cite{OAEI22,he2022machine} on the Unified Medical Language System (UMLS) datasets as part of the Bio-ML track for OAEI 2022. The Ontology Alignment Evaluation Initiative (OAEI) organizes yearly campaigns on ontology matching tasks. Our solution surpasses state-of-the-art LogMap, AML models,  Edit-similarity, and recently proposed BERTMap, AMD, LogMap-Lite, BERTMap-Lite, LSMatch, Matcha and ATMatcher, while offering log-linear complexity in contrast to quadratic in many existing approaches. 

The remainder of this paper is as follows. Section \ref{Relwork} reviews the recent SOTA-related works on OM/OA; Section \ref{Methodology} defines the problem statement, provides a high-level understanding of our proposed approach and the ontologies used; Section \ref{TM} describes TM in detail, elaborates on pre-training, fine-tuning, zero-shot learning, and predictions; Section \ref{TM:results} shows the evaluation criteria, results, and gives insight about the overall model performance; and lastly, Section \ref{Conc} provides a detailed discussion, conclusion on the framework, and outlines our potential future works.

\section{Related Work}\label{Relwork}

OM classical approaches are primarily based on non-contextual matching. Related to that, some notable works in the field of OM include Edit-Similarity \cite{deeponto}, LSMatch \cite{sharma2022lsmatch}, LogMap \cite{jimenez2011logmap}, and AgreementMakerLight (AML) \cite{faria2013agreementmakerlight}, among others. Edit-Similarity is a naïve lexical matching approach based on normalized edit similarity scores. LSMatch is another lexical matching approach based on string similarity match. LogMap and AML are two classical OM systems with leading performance in many equivalence matching tasks. These two approaches are based on lexical matching, mapping extension {{(adding new mappings for semantically related classes of the current mappings)}}, and mapping repair {{(removing mappings that can lead to logical conflicts)}}. However, these lexical approaches do not consider contextual semantics.

Recently, several OM systems, such as OntoEmma \cite{wang2018ontology}, DeepAlignment \cite{kolyvakis2018deepalignment}, VeeAlign \cite{iyer2020veealign}, leveraged dense word embeddings, in which words are projected into a vector. Word pairs with smaller Euclidean distances in the vector space will have closer semantic meanings. Different techniques are used to generate these embeddings. OntoEmma and \cite{zhang2014ontology} uses word2vec \cite{mikolov2013efficient}, which is trained on Wikipedia; \cite{tounsi2019ontology} uses FastText \cite{bojanowski2017enriching}; LogMap-ML \cite{chen2021augmenting} uses OWL2Vec* \cite{chen2021owl2vec}, which is a word2vec model trained on corpora extracted from the ontology with different kinds of semantics; DeepAlignment uses refined word embeddings using counter-fitting; VeeAlign proposes dual embeddings using class labels. These are primarily traditional non-contextual word embedding methods and do not consider word-level contexts. Some of these approaches, such as VeeAlign, are based on supervised training, which requires high-quality labeled mappings for training and can be challenging to obtain.

Recently, transformer-based models~\cite{vaswani2017attention}, thanks to their ability to learn textual contexts, obtained SOTA for several tasks in natural language processing such as machine translation \cite{johnson2017google,xu2021editor,liu2021re}, question answering \cite{clark2019boolq}, among others. Similarly, in the field of OM, recent developments have also shown the potential of using transformer-based frameworks \cite{neutel2021towards,he2022bertmap,wang2022amd}.  Neutel and de Boer (\citeyear{neutel2021towards}) employed contextual BERT embeddings to match two domain ontologies associated with occupations. Each sentence is embedded using BERT, and similarity is applied to get the scores for OM. More recently, \cite{he2022bertmap} proposed BERTMap model, which is obtained by fine-tuning the already pre-trained BERT model for the binary classification task. The BERTMap model often outperformed non-contextual approaches such as LogMap, AML, and LogMap-ML. AMD \cite{wang2022amd} is another recent context-based matching approach that uses a BERT-based model to generate mappings and then filters these mappings using graph embedding techniques. Other related ontology matching systems that participated in OAEI 2022 \cite{OAEI22} are LogMap-Lite, BERTMap-Lite, Matcha, and ATMatcher.

\section{Methodology}\label{Methodology}

\subsection{Problem statement}\label{Methodology:ProbState}
Ontology Matching (OM) or Ontology Alignment (OA) is the process of finding correspondence between the entities/classes of two ontologies \cite{he2022machine}. In this work, a new perspective is presented by treating OM as a translation task for equivalence matching and can be mathematically presented as $f(c_1, T)$, where function $f$ gives the matching target ontology class $c_2 \in {C_2}$, given a source class $c_1 \in {C_1}$, and $T$ is the alignment task identifier. ${O_1}$ and ${O_2}$ as the source and target ontologies, with ${C_1}$ and ${C_2}$ being their respective named class sets. Since we are training a multi-task model, a unique identifier is used for each task.

\begin{figure*}[h]
\centering\includegraphics[width=12cm]{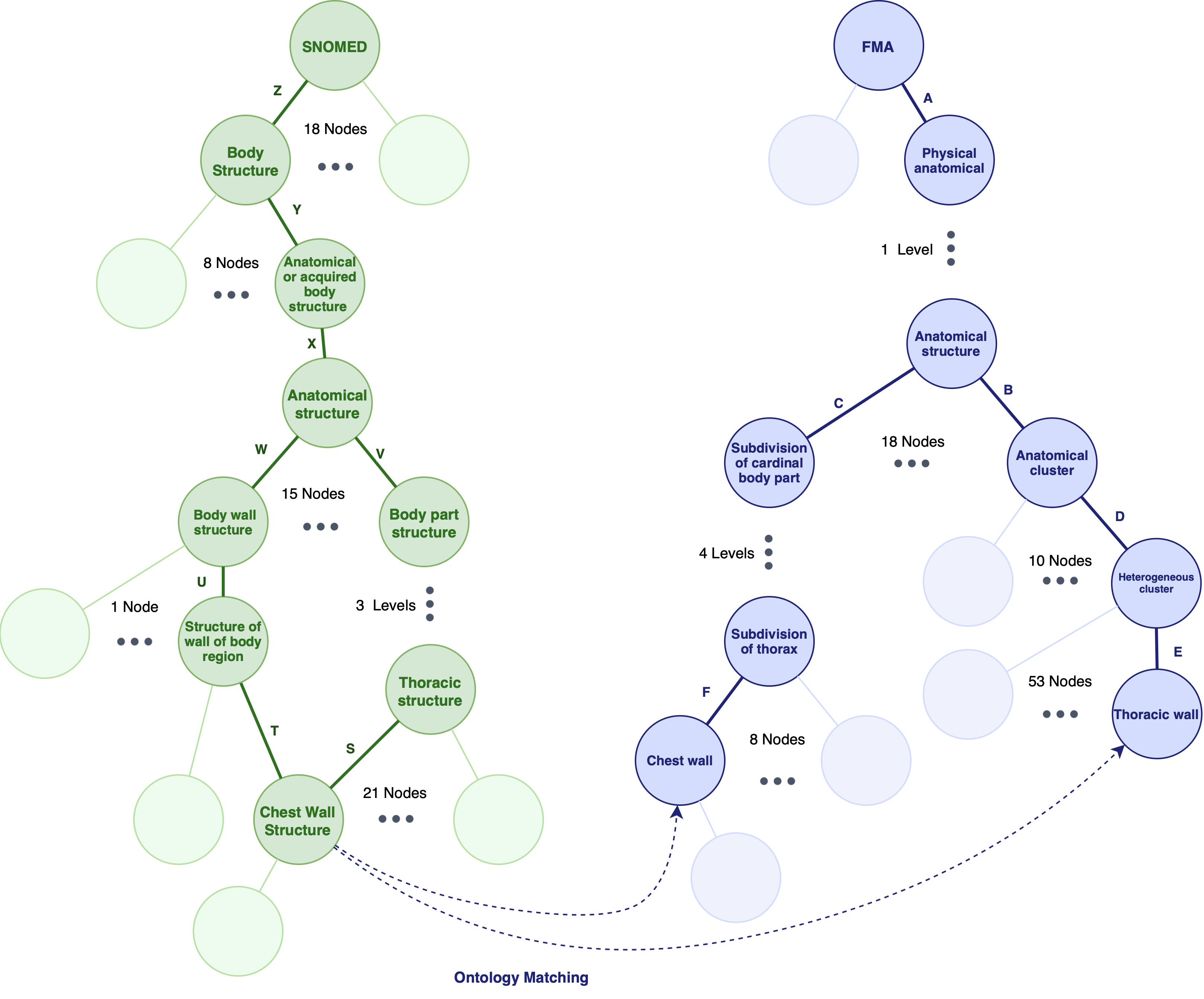}\caption{The equivalence matching between the SNOMED class ID 78904004 – ``Chest Wall Structure'' and two FMA concepts, ``Wall of thorax'' with ID of fma10428 and ``Chest wall'' with ID of fma50060, is illustrated in this figure. TM translates from the source node encoding ``Chest Wall Structure'' in the SNOMED graph to the highlighted path ``A \ldots C \ldots F'' (presenting Chest Wall) and ``A \ldots B \ldots E'' (Thoracic Wall) in FMA ontology. While the SNOMED graph's ``Chest Wall Structure'' node and the FMA graph's ``Chest Wall'' node have children, the FMA ontology's ``Thoracic Wall'' is considered a leaf in this graph (no children).}
\label{fig1}
\end{figure*}

The present work focuses on equivalence matching, where classes having the same semantic meaning in different ontologies are matched with each other. As shown in Figure \ref{fig1}, each ontology is presented in the form of a hierarchical graph structure with parent-child relation, where each class presents a node in the given ontology graph. In Figure \ref{fig1}, we illustrate our high-level solution, where we train our model to learn this hierarchical structure, and consequently, target class $c_2 \in {C_2}$ is obtained as a path in the target ontology graph, for a given input node representing class $c_1 \in {C_1}$ in the input ontology\footnote{Note, each class is presented as a node in the ontology hierarchical graph-structure, as such, class and node are used interchangeably, as appropriate.}.

\subsection{Ontologies}\label{Methodology:Ontologies}

In this work, as a part of the Bio-ML track \cite{OAEI22}, we focus on three UMLS equivalence matching tasks, SNOMED to FMA (Body), SNOMED to NCIT (Neoplas), and SNOMED to NCIT (Pharm), in an unsupervised setting from \cite{OAEI22}, where the matching pairs between these ontologies are only divided into validation (10\%) and testing (90\%) sets, without any training data. Pharm, Neoplas, and Body are associated with the semantic types of ``Pharmacologic Substance'', ``Neoplastic Process'', and ``Body Part, Organ, or Organ Components'' in UMLS, respectively. Based on these semantics types, subset ontologies are provided in \cite{OAEI22}, and are given as SNOMED (Body), SNOMED (Neoplas), SNOMED (Pharm), FMA (Body), NCIT (Neoplas) and NCIT (Pharm), where the first three are the source and last three are the target ontologies in our matching task (Table \ref{tab1}). For each of the classes present in the given ontologies, class ID is provided along with its associated label and possible synonyms (class descriptions). For example, in Figure \ref{fig1}, for Snomed ID 78904004, the class label is “Chest Wall Structure,” and its synonyms are ``Thoracic Wall'' and ``Chest Wall''. 

\begin{table}
\centering
\begin{tabular}{llll} 
\hline
Ontologies & \#Classes & Subsets & \#Classes\\
\hline
\multirow{2}{*}{SNOMED} & \multirow{2}{*}{358,222} & Body &  24,182 \\ 
        &         & Pharm &  16,045\\ 
                   &         & Neoplas &  11,271  \\ 
\hline
FMA   & 104,523   & Body &  64,726\\ 
\hline
\multirow{2}{*}{NCIT}  & \multirow{2}{*}{163,842}  & Pharm &  15,250\\ 
  &      & Neoplas &  13,956 \\
\hline
\end{tabular}
\caption[tab1]{Ontologies and their subsets \cite{OAEI22}, same version as \cite{he2022machine}. SNOMED subsets are the source ontologies, while FMA and NCIT are the target ontologies.}
\label{tab1}
\end{table}

\begin{figure*}[h]
\includegraphics[clip, trim={0cm 0cm 0cm 3cm},scale=.5]{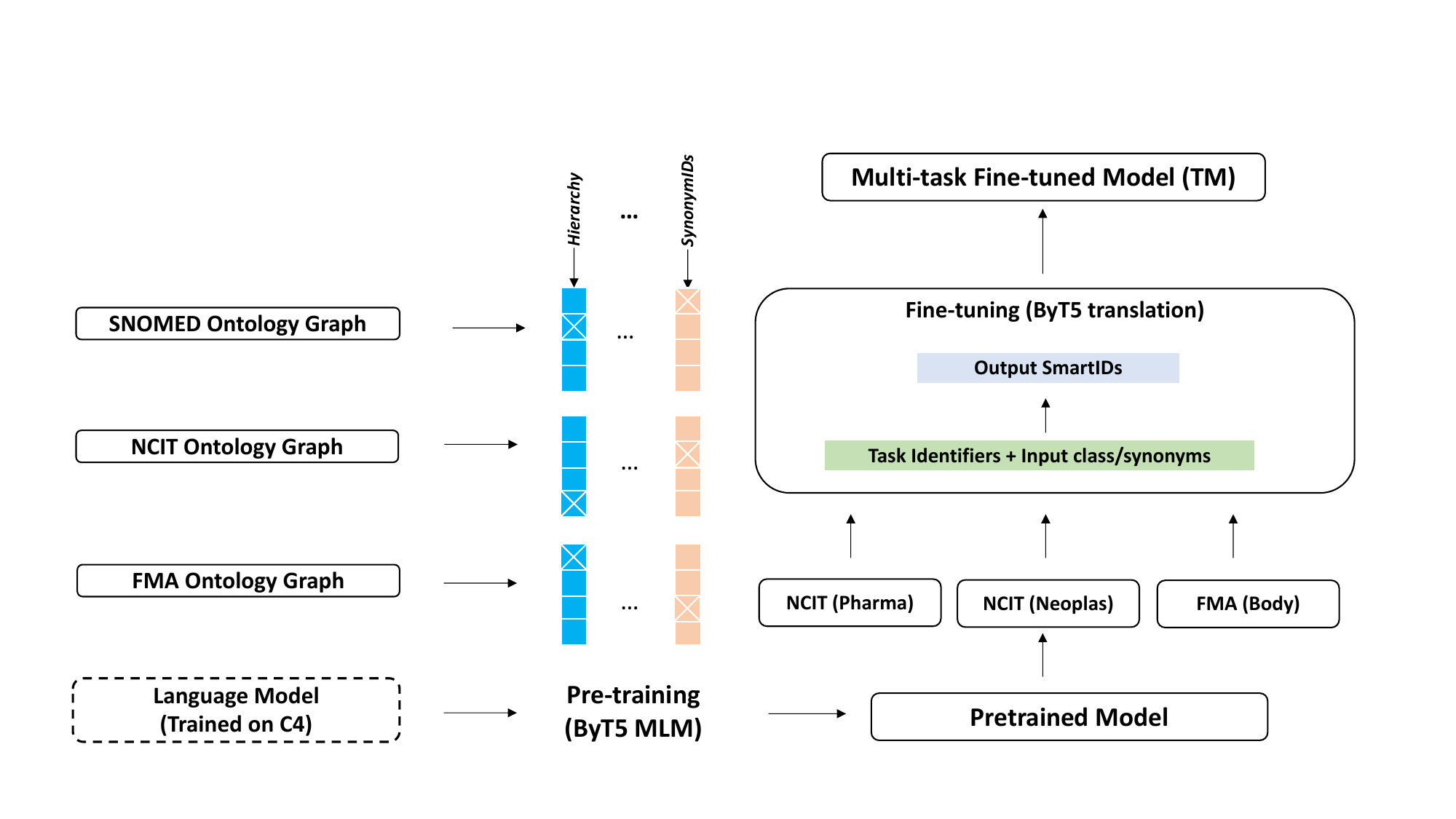}
\caption{Training Architecture. Starting from a language model pre-trained on the C4 dataset, further pre-training is done using MLM on the full ontology graphs. The pre-trained model is then fine-tuned on downstream tasks, translating from the class descriptions (label and synonyms) to the target node path (hierarchical-IDs). The pre-training and fine-tuning are done in a multi-task manner.  The pre-training is performed on both source and target inner-ontologies, and fine-tuning is done on task specific target subset ontologies.}\label{fig: train_arch}
\end{figure*}

\begin{figure}[h]
\includegraphics[clip, trim={-3cm 0cm 0cm 0cm},width=7cm]{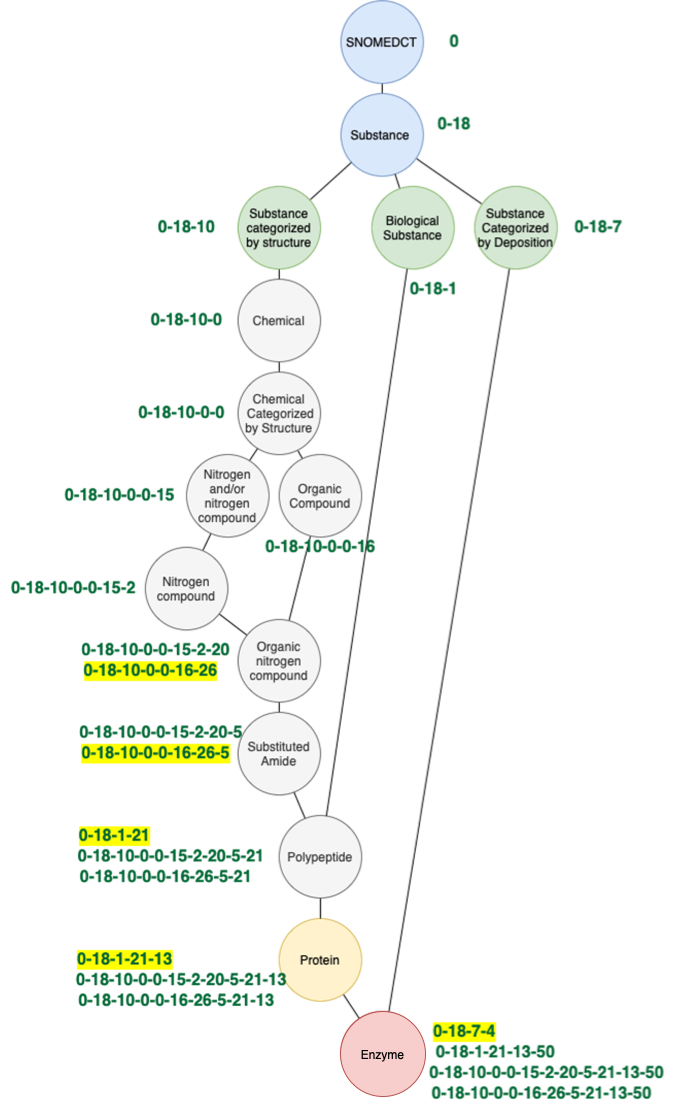}
\caption{Hierarchical-IDs generation. This diagram illustrates hierarchical-IDs generation for the Enzyme concept in the SNOMED ontology. The enzyme has four paths because this node has multiple parents. The shortest ID (highlighted) is chosen as a Hierarchical-ID, and others are SynonymIDs for this concept.}\label{fig: smartids}
\end{figure}

\section{Truveta Mapper (TM): Proposed approach for OM}\label{TM}

Figure \ref{fig: train_arch} demonstrates training architecture, with two main steps of pre-training and fine-tuning.
 Starting from a language model pre-trained on the C4 dataset, the model is further pre-trained on the full ontologies, learning each ontology’s semantics and hierarchical structure. Afterward, the model is trained on the downstream task using the subset ontology data during the fine-tuning stage. The pre-training and fine-tuning steps are done in a multi-task manner on inner-ontologies, which enables the model in extensive transfer-learning (Figure \ref{fig: train_arch}). In the prediction step, given a source ontology, the output is predicted in a zero-shot manner. More details are provided for each step in the subsequent subsections.

\subsection{Pre-training}\label{TM:pre-training}

\paragraph{Hierarchical-ID generation.} An ontology is represented in the form of a graph where each node represents a class, and the parent and child relations of the ontology serve as connections between classes.
Based on this graph structure of each full ontology, hierarchical-IDs are generated for all the classes. These are constructed by starting from the root node, separated by ``-'' at each hierarchy level, and traversing through each node in that level as shown in Figure \ref{fig: smartids}. Following this method, a unique ID is generated for each path traversed. As such, for ontologies like SNOMED, where there are multiple paths between the root and any given class, there could be multiple IDs for that node. In such cases, the shortest ID is considered the hierarchical-ID of that node (highlighted in yellow in Figure \ref{fig: smartids}), while the other path IDs are considered its synonymIDs. Each node ID inherently captures the information of all its ancestors. This enables the model to trace from a broader class, starting from the root and getting more granular at each level, thus simplifying the translation task.

\paragraph{Training.} 



After generating the hierarchical-IDs, multi-task pre-training is done on the full ontologies {{(SNOMED, FMA, NCIT)}} using MLM by randomly masking the nodes, enabling the model to learn the hierarchy and semantics. For instance,
``Structure of Forel's H2 bundle'' is represented as ``1-1-0-0-0-0-4-1-1-0-0-0-7'' and is masked as ``1-1-0-0-0-0-[MASK]-1-0-0-0-7''. Furthermore, additional tasks are included in order for the model to learn the semantics of each class in the form of class-level synonyms, labels, and descriptions; class-level relations between child and parent nodes; and the relation between synonym-ID and hierarchical-ID, using separate identifiers for each task in the pre-training step (Figure \ref{fig: train_arch}). {{
Task identifiers are added in the form of prefixes, to distinguish between different ontologies. For example, SNOMED ontology is prefixed as ``F0:'', where ``F'' represents fully specified name and ``0'' indicates SNOMED Ontology. Similarly, FMA and NCIT are represented using ``1'' and ``2'' identifiers. Some representative examples are presented in Table~\ref{tab:pre-training}, where similar tasks are defined for each ontology, with the objective of learning the hierarchical structure and semantics of each using MLM.}}
{{Based on the tasks stated in Table \ref{tab:pre-training}, we generate the pre-training dataset which}} has 2,406,456 instances constituting SNOMED, NCIT, and FMA ontologies. The model is trained for 3 epochs, with an increasing masking percentage linearly over time, starting at 10\% and increasing to 35\% in the final batch. The pre-training is done on 8 V100 32GB Nvidia GPUs with a batch size of 20, using a learning rate of 1e-3 with linear decay scheduler and AdamW optimizer. 

\begin{table*}
\centering
\begin{tabular}{ll} 
\hline
Pre-training tasks & Example \\
\hline
 ID: Child \& Parent &  \textbf{CP:} \textbf{0-}2-3-1-43-1 \textbf{0-}2-3-42-1\\ 
 ID: SYN \& SYN    & \textbf{0-}2-3-1-43-1 \textbf{0-}2-3-42-1-0\\ 
\hline
ID \& SYN       & \textbf{0-}10-40-13-9 \textbf{S0:} Oromucosal solution for gargle \\ 
ID \& FSN       & \textbf{0-}9-18-6-10-28-2 \textbf{F0:} Coagulation or electrocoagulation of inner ear\\ 
\hline
FSN \& SYN & \textbf{F0:} Coagulation or electrocoagulation of inner ear \textbf{S0:} electrocoagulation of ear \\ 
\hline
\hline
Fine-tuning tasks & Example (Source ontology FSN/SYN input, Target ontology ID output) \\
\hline
Pharm & F0: Ceprotin \textbf{2-}, 5-5-6-4-93 \\ 
Neoplas  & F0: FAB M1 \textbf{2-}, 0-0-6-2-1-2-8-0-0-7 \\ 
Body & F0: Hairs set \textbf{1-}, 1-0-0-1-7-0-41 \\ 
\hline
\end{tabular}
\caption[tab:pre-training]{{Pre-training and fine-tuning tasks examples. Task identifiers are in bold. During pre-training, a random part of the input in the examples get masked. During fine-tuning root node of the hierarchical-ID is used as a suffix to provide appropriate task identifier. Different identifiers are used for FMA and NCIT, and acronyms SYN and FSN corresponds to synonyms and Fully-Specified-Name respectively.}} 
\label{tab:pre-training}
\end{table*}


In this work, ByT5 \cite{xue2022byt5}, which is a token-free variation of mT5 \cite{xue2020mt5} and supports multi-task training, is used as the model structure for pre-training, fine-tuning and zero-shot predictions.

\subsection{Fine-tuning}\label{TM:fine-tuning}

The fine-tuning step aims to train the model on the downstream OM tasks. Only target subset ontologies, i.e., NCIT (Pharm), NCIT (Neoplas), and FMA (Body), are used for fine-tuning. The training data of each target sub-ontologies is augmented using the exact matches present in the labels and synonyms of other subset ontologies. We are also taking advantage of older ontology versions to add more synonyms to each target label. This expands the training corpus, enriches the data with minimal processing, and helps to perform more comprehensive learning. After the data augmentation for all the target sub-ontologies, fine-tuning is performed only on these target sub-ontologies corpora, i.e., NCIT (Pharm), NCIT (Neoplas), and FMA (Body).

\begin{figure*}[h]
\includegraphics[clip, trim={4cm 6cm 7cm 5cm}, scale=0.7]{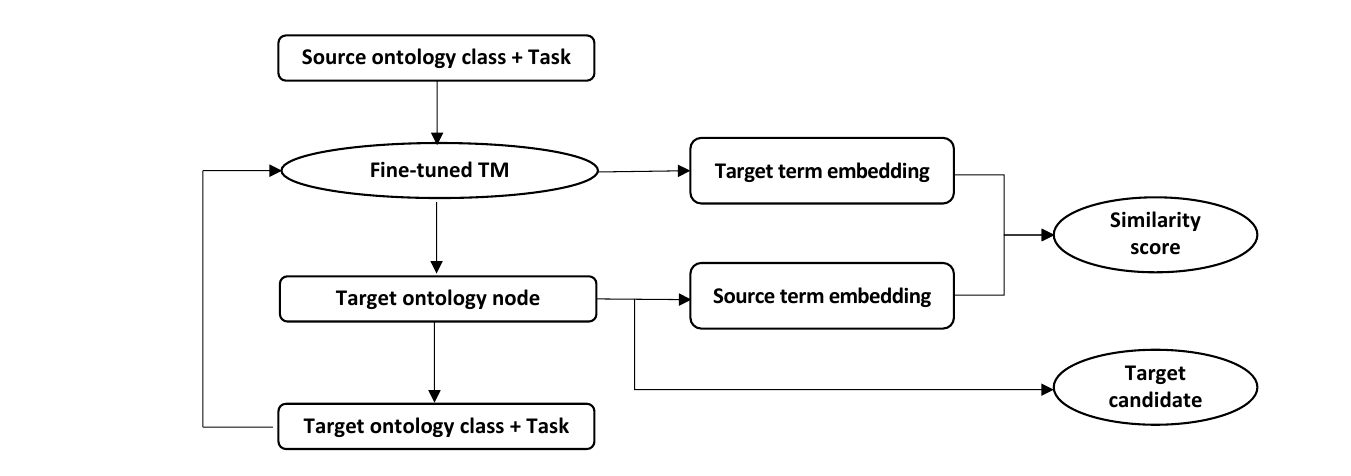}
\caption{Zero-shot predictions. Given a source term and the assigned translation task (e.g.,  SNOMED to FMA), the output is generated in two steps: Prediction step and Validation step. In the Prediction step, a potential target candidate is generated along with the embeddings associated with the source term. In the Validation step, the target candidate class is again passed through our translation model to generate embeddings. Based on the source and target terms embeddings, a similarity score between the source and target candidate is obtained. This is done in a zero-shot manner with time complexity of $O(log(n))$.}\label{fig: pred}
\end{figure*}

Training data is generated for each class in the target ontologies, where the input is the class label, synonyms, and descriptions, and output is the corresponding node hierarchical-ID, using a separate identifier for each task. {{Similar to our pre-training approach, multi-task fine-tuning is performed on the downstream OM tasks. Some examples are presented in Table~\ref{tab:pre-training}.}}

{{Based on the fine-tuning tasks described in Table \ref{tab:pre-training}, we generate the fine-tuning training data which has 462,789 samples from}} Pharm, Neoplas, and Body subsets. Using 8 Nvidia V100 32GB GPUs with a batch size of 20, the fine-tuning took around 21 epochs. For the fine-tuning, a learning rate of 1e-3 with linear decay scheduler and warm-up of 1.5 epoch using AdamW optimizer with eps of 1e-8 and weight decay of 1e-2 is used. 

\subsection{Zero-shot Predictions}\label{TM:predictions}

TM is a multi-task model with the capability to translate between multiple ontologies from the input source class labels/synonyms to target hierarchical-IDs. Thus, given a source term, the model predicts the potential candidate in the target ontology graph. For confidence scoring, two approaches have been adopted here: 
(i) Greedy search score: Scores are generated based on greedy search with softmax probabilities using temperature scaling. This is a naive way to compute the confidence directly from the model prediction.
(ii) Using embeddings: This is a sophisticated method proposed to make the TM predictions more robust and improve model precision, by leveraging semantic similarity using embeddings of source terms and predicted target candidates. Using the same model, the embeddings are generated for the target candidate and the similarity score is obtained between the source term and predicted target term embeddings.  (Figure \ref{fig: pred}). Scores are generated across the source and predicted class labels and synonyms, all of which are also augmented by singularization. The maximum generated score is considered as the similarity score. As such, the proposed model takes advantage of both graph search and semantic matching. Mathematically, similarity score $S$ is given as:
\begin{equation}
S =  
\begin{cases}
    \ 1.0, \;\;\;\;\;\;\;\;\;\;\;\;\;\;\;\;\;\;\;\;\;\;\;\;\;\;\;\;\;\;\;\; \text{if } \Omega(c_1) \cap \Omega(c_2) \neq \emptyset &  \\
     max(Sim(\Omega(c_1),\Omega(c_2)), \;\text{otherwise} &
\end{cases}
\end{equation}
where $c_2$ is the predicted class for $c_1$, $\Omega(c_1)$ and $\Omega(c_2)$ are sets of labels and synonyms for $c_1$ and $c_2$, respectively, and 
$max(Sim(\Omega(c_1),\Omega(c_2))$ selects the maximum cosine similarity score across all the labels and synonyms of $c_1$ (source) and $c_2$ (predicted). If an exact match is available between the labels and synonyms of source and target classes, we assign a maximum similarity score, since embedding similarity will also give a similar result. The source and the target candidates are considered valid mapping pairs if their similarity score exceeds a selected threshold for both the approaches.

{{One of the main advantages of our proposed TM is that it reduces the time complexity to log-linear as opposed to the naive solution of search that results in quadratic complexity\footnote{Note that BERTMap reduces the time complexity from $O(n^2)$ in traditional approaches to $O(kn)$, where $k<<n$ with an additional preprocessing step by considering only a small portion of target subset ontology classes with at least one subword token common to the source class candidate, which adds dependency on the tokenization hyperparameters and could be error prone since some semantically matching cases with lexical variations could get filtered out in this process. 
Contrary to that, such limitation does not exist in TM since it performs matching from source to target without reducing the target corpora size.}. Given an input term with a specified task identifier, TM is able to predict the best possible match from the target ontology with $O(log(n))$ complexity, where $n$ corresponds to the number of nodes in the target ontology graph (same as the number of classes). Overall, TM reduces the time-complexity to $O(nlog(n))$, noting that a single search in a tree structure with $n$ nodes can be performed in $O(log(n))$ time.}}

\section{Results}\label{TM:results}

\subsection{Evaluation criteria}

Commonly used metrics for evaluating OM systems \cite{he2022machine}: Precision (P), Recall (R), and F-score are used as the global evaluation metrics. Mathematically,
\begin{equation}
\begin{aligned}
& P=\frac{\left|  M_{out}  \cap M_{ref} \right|}{\left| M_{out}\right|}  \; , \;\;\;\;\  R=\frac{\left|  M_{out}  \cap M_{ref} \right|}{\left| M_{ref}\right|}  \\
& F_{\beta} = (1+\beta^2)\frac{P.R}{\beta^2.P+R}
\end{aligned}
\label{Eq:global}
\end{equation}
where, $M_{ref}$ are the reference mappings, consisting of matching pairs, $m=(c,c')$, such that $c$ and $c'$ are two classes from the to-be-aligned ontologies, and  $M_{out}$ are the mappings computed by OM systems and $\beta=1$.

Local evaluation metrics,  $Hits@K$  and Mean Reciprocal Rank ($MRR$),  introduced in \cite{he2022machine} are also used for current evaluation and can be represented as:  
\begin{equation}
\begin{aligned}
&Hits@K=\frac{\left| \{m \in  M_{ref} | Rank(m) \leq K \} \right|}{\left| M_{ref}\right|} \\
&MRR=\frac{ \sum_{m \in  M_{ref}} Rank(m)^{-1} }{\left| M_{ref}\right|}
\end{aligned}
\label{Eq:global}
\end{equation}
where $Rank(m)$ returns the ranking position of $m$ among $M_m \cup \{m\}$ according to their scores,  $M_m$ represents a set of negative mappings pairs for each of the source term $c$ in $M_{ref}$, such that $(c,c''_i) \in M_m$ with $i \in \{1,2,...,100\}$ and  $c''_i$ are the 100 negative output candidates from target ontologies for each of the source terms $c$ in $M_{ref}$. As such, the Hits and MRR would be different for different selected 100 samples. We have published the results of our model based on the provided $M_m$ set in \cite{he2022machine} for a fair comparison. To provide a more robust measure of local metrics, we are reporting overall accuracy as well, although this is not provided for any of the other models. Accuracy here can be mathematically presented as:
\begin{equation}
Accuracy=\frac{\left| \{m \in  M_{ref} | f(c,T) = c' \} \right|}{\left| M_{ref}\right|} 
\end{equation}\label{Eq:local_new}
where $m=(c,c')$ represents matching pairs in the $M_{ref}$ set, and $f(c,T)$ refers to the target candidate predicted by the model,  given an input term $c$ and appropriate task identifier $T$.

\paragraph{Baselines. } Results are compared with the SOTA approaches: Edit-Similarity, LogMap, AML, BERTMap \cite{he2022machine}, and recently published results in \cite{OAEI22}. To be consistent, evaluation for P, R, F-score, Hit@1, and MRR is done using \cite{deeponto} library.

\subsection{{Prediction Results}}

Prediction results are shown in Tables \ref{tab:res1}--\ref{tab:res3}, for the three equivalence OM tasks, from SNOMED to FMA (Body), SNOMED to NCIT (Pharm), and SNOMED to NCIT (Neoplas).  The results demonstrate precision, recall, F-score, Hit@1, MRR, and accuracy for TM and baseline approaches presented in \cite{he2022machine} and \cite{OAEI22} on the test data for the unsupervised setting. In the given tables, superscripts$^{1,2}$ are based on our proposed TM, where the former is based on embedding similarity score and later is based on greedy search score, superscript$^*$ results are based on \cite{he2022machine}, and superscript$^{**}$ correspond to \cite{OAEI22} published results. The highest numbers for each of these metrics are highlighted in the tables to emphasize which model is outperforming others in each category.

The overall results illustrate that TM is outperforming all the baselines for all three OM tasks in F-score, Hit@1, and MRR. A high threshold is selected to generate the most confident cross-ontology matching pairs. Note that a single unified model is trained and leveraged here to predict all the results in the form of a source class to target hierarchical-IDs, using appropriate task identifiers.

\begin{table*}
\centering
\begin{tabular}{lllllll} 
\hline
Task & Precision & Recall & F-score & MRR & Hit@1 &Accuracy\\
\hline
TM(Ours)$^1$ &  0.947	& 0.738	& \textbf{0.830}	& \multirow{2}{*}{\textbf{0.960}}	&  \multirow{2}{*}{\textbf{0.942}}  &\multirow{2}{*}{\textbf{0.801}}\\
TM(Ours)$^2$  &  0.960	& 0.720	& 0.823	& 	&    &\\ 
\hline
Edit-Similarity$^*$ 	&0.976	&0.660	&0.787	&0.895    &0.869   &NA\\ 
LogMap$^*$	&0.702	&0.581	&0.636	&0.545    &0.330 &NA\\ 
AML$^*$	&0.841	&\textbf{0.776}	&0.807	&NA	         &NA &NA\\ 
BERTMap$^*$&\textbf{0.997}	&0.639	&0.773	& 0.954  &0.930 &NA\\ 
LogMap-Lite$^{**}$	&0.967	&0.543	&0.695	&NA	&NA&NA	\\		
AMD	 $^{**}$                 &0.890	&0.704	&0.786	&NA	&NA&NA	\\	
BERTMap-Lite$^{**}$	&0.976	&0.660	&0.787	&0.895	&0.869	&NA\\
Matcha$^{**}$	&0.875	&0.594	&0.707		&NA	&NA	&NA\\	
ATMatcher	$^{**}$&0.264	&0.226	&0.244	&NA	&NA	&NA\\	
LSMatch$^{**}$	&0.809	&0.072	&0.132	&NA	&NA	&NA\\	
\hline	
\end{tabular}
\caption{Result for equivalence matching – SNOMED (Body) to FMA (Body).}
\label{tab:res1}
\end{table*}

\begin{table*}
\centering
\begin{tabular}{lllllll} 
\hline
Task & Precision & Recall & F-score & MRR & Hit@1 &Accuracy\\
\hline
TM(Ours)$^1$&  0.972	& \textbf{0.929}	& \textbf{0.950}	& \multirow{2}{*}{\textbf{0.987}}	& \multirow{2}{*}{\textbf{0.982}}  &  \multirow{2}{*}{\textbf{0.946}}\\ 
TM(Ours)$^2$ &  0.977	& 0.872	& 0.922	& 	&   & \\ 
\hline
Edit-Similarity$^*$ 	&0.979 &0.432 &0.600 &0.836 &0.760   &NA\\ 
LogMap$^*$	&0.915 &0.612& 0.733 &0.820& 0.695&NA \\ 
AML$^*$	&0.940 &0.615 &0.743	&NA	         &NA &NA\\ 
BERTMap$^*$&0.966 &0.606 &0.745 &0.919 &0.876 &NA\\ 
LogMap-Lite$^{**}$	&\textbf{0.995}	& 0.598	& 0.747	&NA	&NA	&NA\\		
AMD	$^{**}$                 &0.962	& 0.745	& 0.840	&NA	&NA	&NA\\	
BERTMap-Lite$^{**}$	&0.979	& 0.432	& 0.600	& 0.836	& 0.760	&NA\\
Matcha$^{**}$	&0.941	& 0.613	& 0.742		&NA	&NA	&NA\\	
ATMatcher	$^{**}$&0.937	& 0.566	& 0.706	&NA	&NA	&NA\\	
LSMatch$^{**}$	&0.982	& 0.551	& 0.706	&NA	&NA	&NA\\	
\hline	
\end{tabular}
\caption{Results for equivalence matching – SNOMED (Pharm) to NCIT (Pharm).}
\label{tab:res2}
\end{table*}

\begin{table*}
\centering
\begin{tabular}{lllllll} 
\hline
Task & Precision & Recall & F-score & MRR & Hit@1 &Accuracy\\
\hline
TM(Ours)$^1$& 0.809 	& \textbf{0.795}	& \textbf{0.802}	& \multirow{2}{*}{\textbf{0.962}}	& \multirow{2}{*}{\textbf{0.944}}  & \multirow{2}{*}{\textbf{0.802}}\\ 
TM(Ours)$^2$& 0.812 	& 0.773	& 0.792	& 	&   & \\ 
\hline
Edit-Similarity$^*$ 	&0.815 & 0.709 & 0.759 & 0.900 & 0.876 &NA \\ 
LogMap$^*$	&0.823 & 0.547 & 0.657 & 0.824 & 0.747&NA \\ 
AML$^*$	&0.747 & 0.554 & 0.636	&NA	         &NA&NA \\ 
BERTMap$^*$&0.655 & 0.777 & 0.711 & 0.960 & 0.939&NA\\ 
LogMap-Lite$^{**}$	&\textbf{0.947}	& 0.520	& 0.671	&NA	&NA&NA	\\		
AMD	 $^{**}$                 &0.836	& 0.534	& 0.652	&NA	&NA	&NA\\	
BERTMap-Lite$^{**}$	&0.815	& 0.709	& 0.759	& 0.900	& 0.876&NA	\\
Matcha$^{**}$	&0.754	& 0.564	& 0.645		&NA	&NA	&NA\\	
ATMatcher	$^{**}$&0.866& 	0.284	& 0.428	&NA	&NA	&NA\\	
LSMatch$^{**}$	&0.902	& 0.238	& 0.377	&NA	&NA&NA	\\	
\hline
\end{tabular}
\caption{Result for equivalence matching – SNOMED (Neoplas) to NCIT (Neoplas).}
\label{tab:res3}
\end{table*}

There are two TM results presented in the given tables, and both are based on different scoring schemes. TM$^2$ is based on greedy search scores and TM$^1$ is based on a new and more robust prediction scheme {{using embeddings}} described in Subsection \ref{TM:predictions}, taking advantage of both graph search and semantic similarity. It can be seen that both of our methods surpass SOTA for all the tasks, but TM$^1$ is more robust and has significant improvements as compared to any of the existing methods. To be precise, 2.3\% improvement over the second best result (AML) in Body, 11.0\% improvement for Pharm (as compared to AMD), and 4.3\% improvement for Neoplas as compared to BertMap-Lite and Edit-Similarity, is seen for TM$^1$ in the F-score. It should be noted that even without TM, none of these methods are SOTA in all the tasks.

For generating local metrics for Hit@1 and MRR, TM is used to generate the embedding similarity score of input terms in the test set and their corresponding candidates in $M_m \cup \{m\}$ set. We are also outperforming all existing SOTA methods  based on MRR and Hit@1. Additionally, we are reporting accuracy metric, which is consistent, and more representative of the model performance. For this metric, the TM predictions are obtained across the entire target ontology without using any smaller subset of negative samples from the test set, while reducing the time complexity from quadratic to log-linear.

\section{Conclusions and Discussions}\label{Conc}
This work presents a new approach to OM by treating the OM process as a translation task and performing multi-task pre-training, fine-tuning, and predictions in a zero-shot, unified and end-to-end manner. The proposed approach takes advantage of transfer learning across different ontologies and does not require manual annotations for training. Additionally, the trained model understands the semantics of the text as well as the structure of the ontologies. We show that our proposed method outperforms Edit-Similarity, LogMap, AML, BERTMap, and the recently proposed OM frameworks in the OM22 conference \cite{OAEI22} in all the tasks.  

Our approach provides several advantages: (1) It reduces the time complexity to log-linear during inference, (2) It is based on zero-shot prediction, without requiring much post-processing and does not employ mapping extension or mapping repair in contrast to the other methods, (3) It does not require any manual labeled cross-ontologies matching pairs due to zero-shot learning, (4) One unified framework is used as a result of multi-tasking, which makes it easier to productionize these large transformer-based models, (5) It is robust toward different tokenization schemes as it uses byte level tokenization, (6) It learns complete ontologies graphs, using the hierarchical-IDs which provides a more natural path for translation, and would be significantly helpful for subsumption mappings. 

In the future, we will pre-train the starting checkpoint with more domain-related corpus (e.g., PubMed, MIMIC-III, clinical notes) instead of the C4 dataset. Another interesting track can be ensemble learning of existing SOTA models with TM.



\bibliographystyle{named}
\bibliography{ijcai23}

\begin{thebibliography}{}

\bibitem[\protect\citeauthoryear{Bojanowski \bgroup \em et al.\egroup
  }{2017}]{bojanowski2017enriching}
Piotr Bojanowski, Edouard Grave, Armand Joulin, and Tomas Mikolov.
\newblock {Enriching word vectors with subword information}.
\newblock {\em Transactions of the association for computational linguistics},
  5:135--146, 2017.

\bibitem[\protect\citeauthoryear{Chen \bgroup \em et al.\egroup
  }{2021a}]{chen2021owl2vec}
Jiaoyan Chen, Pan Hu, Ernesto Jimenez-Ruiz, Ole~Magnus Holter, Denvar
  Antonyrajah, and Ian Horrocks.
\newblock {Owl2vec*: Embedding of owl ontologies}.
\newblock {\em Machine Learning}, 110(7):1813--1845, 2021.

\bibitem[\protect\citeauthoryear{Chen \bgroup \em et al.\egroup
  }{2021b}]{chen2021augmenting}
Jiaoyan Chen, Ernesto Jim{\'e}nez-Ruiz, Ian Horrocks, Denvar Antonyrajah, Ali
  Hadian, and Jaehun Lee.
\newblock {Augmenting ontology alignment by semantic embedding and distant
  supervision}.
\newblock In {\em European Semantic Web Conference}, pages 392--408. Springer,
  2021.

\bibitem[\protect\citeauthoryear{Clark \bgroup \em et al.\egroup
  }{2019}]{clark2019boolq}
Christopher Clark, Kenton Lee, Ming-Wei Chang, Tom Kwiatkowski, Michael
  Collins, and Kristina Toutanova.
\newblock {BoolQ: Exploring the surprising difficulty of natural yes/no
  questions}.
\newblock {\em arXiv preprint arXiv:1905.10044}, 2019.

\bibitem[\protect\citeauthoryear{{DeepOnto}}{2022}]{deeponto}
{DeepOnto}.
\newblock \url{https://github.com/KRR-Oxford/DeepOnto}.
\newblock Accessed: 2022-10-14, 2022.

\bibitem[\protect\citeauthoryear{Devlin \bgroup \em et al.\egroup
  }{2018}]{devlin2018bert}
Jacob Devlin, Ming-Wei Chang, Kenton Lee, and Kristina Toutanova.
\newblock {Bert: Pre-training of deep bidirectional transformers for language
  understanding}.
\newblock {\em arXiv preprint arXiv:1810.04805}, 2018.

\bibitem[\protect\citeauthoryear{Faria \bgroup \em et al.\egroup
  }{2013}]{faria2013agreementmakerlight}
Daniel Faria, Catia Pesquita, Emanuel Santos, Matteo Palmonari, Isabel~F Cruz,
  and Francisco~M Couto.
\newblock {The agreementmakerlight ontology matching system}.
\newblock In {\em OTM Confederated International Conferences" On the Move to
  Meaningful Internet Systems"}, pages 527--541. Springer, 2013.

\bibitem[\protect\citeauthoryear{He \bgroup \em et al.\egroup
  }{2022a}]{he2022bertmap}
Yuan He, Jiaoyan Chen, Denvar Antonyrajah, and Ian Horrocks.
\newblock {BERTMap: a BERT-based ontology alignment system}.
\newblock In {\em Proceedings of the AAAI Conference on Artificial
  Intelligence}, volume~36, pages 5684--5691, 2022.

\bibitem[\protect\citeauthoryear{He \bgroup \em et al.\egroup
  }{2022b}]{he2022machine}
Yuan He, Jiaoyan Chen, Hang Dong, Ernesto Jim{\'e}nez-Ruiz, Ali Hadian, and Ian
  Horrocks.
\newblock {Machine Learning-Friendly Biomedical Datasets for Equivalence and
  Subsumption Ontology Matching}.
\newblock {\em arXiv preprint arXiv:2205.03447}, 2022.

\bibitem[\protect\citeauthoryear{Iyer \bgroup \em et al.\egroup
  }{2020}]{iyer2020veealign}
Vivek Iyer, Arvind Agarwal, and Harshit Kumar.
\newblock {VeeAlign: a supervised deep learning approach to ontology
  alignment.}
\newblock In {\em OM@ ISWC}, pages 216--224, 2020.

\bibitem[\protect\citeauthoryear{Jim{\'e}nez-Ruiz and
  Cuenca~Grau}{2011}]{jimenez2011logmap}
Ernesto Jim{\'e}nez-Ruiz and Bernardo Cuenca~Grau.
\newblock {Logmap: Logic-based and scalable ontology matching}.
\newblock In {\em International Semantic Web Conference}, pages 273--288.
  Springer, 2011.

\bibitem[\protect\citeauthoryear{Johnson \bgroup \em et al.\egroup
  }{2017}]{johnson2017google}
Melvin Johnson, Mike Schuster, Quoc~V Le, Maxim Krikun, Yonghui Wu, Zhifeng
  Chen, Nikhil Thorat, Fernanda Vi{\'e}gas, Martin Wattenberg, Greg Corrado,
  et~al.
\newblock {Google’s multilingual neural machine translation system: Enabling
  zero-shot translation}.
\newblock {\em Transactions of the Association for Computational Linguistics},
  5:339--351, 2017.

\bibitem[\protect\citeauthoryear{Kolyvakis \bgroup \em et al.\egroup
  }{2018}]{kolyvakis2018deepalignment}
Prodromos Kolyvakis, Alexandros Kalousis, and Dimitris Kiritsis.
\newblock {Deepalignment: Unsupervised ontology matching with refined word
  vectors}.
\newblock In {\em Proceedings of the 2018 Conference of the North American
  Chapter of the Association for Computational Linguistics: Human Language
  Technologies, Volume 1 (Long Papers)}, pages 787--798, 2018.

\bibitem[\protect\citeauthoryear{Liu and Chen}{2021}]{liu2021re}
Huey-Ing Liu and Wei-Lin Chen.
\newblock {Re-transformer: a self-attention based model for machine
  translation}.
\newblock {\em Procedia Computer Science}, 189:3--10, 2021.

\bibitem[\protect\citeauthoryear{Mikolov \bgroup \em et al.\egroup
  }{2013}]{mikolov2013efficient}
Tomas Mikolov, Kai Chen, Greg Corrado, and Jeffrey Dean.
\newblock {Efficient estimation of word representations in vector space}.
\newblock {\em arXiv preprint arXiv:1301.3781}, 2013.

\bibitem[\protect\citeauthoryear{Neutel and de Boer}{2021}]{neutel2021towards}
Sophie Neutel and Maaike~HT de~Boer.
\newblock {Towards Automatic Ontology Alignment using BERT.}
\newblock In {\em AAAI Spring Symposium: Combining Machine Learning with
  Knowledge Engineering}, 2021.

\bibitem[\protect\citeauthoryear{{OAEI}}{2022}]{OAEI22}
{OAEI}.
\newblock \url{https://www.cs.ox.ac.uk/isg/projects/ConCur/oaei/2022/}.
\newblock Accessed: 2022-08-08, 2022.

\bibitem[\protect\citeauthoryear{Sharma \bgroup \em et al.\egroup
  }{2022}]{sharma2022lsmatch}
Abhisek Sharma, Archana Patel, and Sarika Jain.
\newblock {LSMatch and LSMatch-Multilingual Results for OAEI}.
\newblock 2022.

\bibitem[\protect\citeauthoryear{Tounsi~Dhouib \bgroup \em et al.\egroup
  }{2019}]{tounsi2019ontology}
Molka Tounsi~Dhouib, Catherine Faron~Zucker, and Andrea~GB Tettamanzi.
\newblock {An ontology alignment approach combining word embedding and the
  radius measure}.
\newblock In {\em International Conference on Semantic Systems}, pages
  191--197. Springer, Cham, 2019.

\bibitem[\protect\citeauthoryear{Vaswani \bgroup \em et al.\egroup
  }{2017}]{vaswani2017attention}
Ashish Vaswani, Noam Shazeer, Niki Parmar, Jakob Uszkoreit, Llion Jones,
  Aidan~N Gomez, {\L}ukasz Kaiser, and Illia Polosukhin.
\newblock {Attention is all you need}.
\newblock {\em Advances in neural information processing systems}, 30, 2017.

\bibitem[\protect\citeauthoryear{Wang \bgroup \em et al.\egroup
  }{2018}]{wang2018ontology}
Lucy~Lu Wang, Chandra Bhagavatula, Mark Neumann, Kyle Lo, Chris Wilhelm, and
  Waleed Ammar.
\newblock {Ontology alignment in the biomedical domain using entity definitions
  and context}.
\newblock {\em arXiv preprint arXiv:1806.07976}, 2018.

\bibitem[\protect\citeauthoryear{Wang}{2022}]{wang2022amd}
Zhu Wang.
\newblock {AMD Results for OAEI 2022}.
\newblock 2022.

\bibitem[\protect\citeauthoryear{Xu and Carpuat}{2021}]{xu2021editor}
Weijia Xu and Marine Carpuat.
\newblock {Editor: an edit-based transformer with repositioning for neural
  machine translation with soft lexical constraints}.
\newblock {\em Transactions of the Association for Computational Linguistics},
  9:311--328, 2021.

\bibitem[\protect\citeauthoryear{Xue \bgroup \em et al.\egroup
  }{2020}]{xue2020mt5}
Linting Xue, Noah Constant, Adam Roberts, Mihir Kale, Rami Al-Rfou, Aditya
  Siddhant, Aditya Barua, and Colin Raffel.
\newblock {mT5: A massively multilingual pre-trained text-to-text transformer}.
\newblock {\em arXiv preprint arXiv:2010.11934}, 2020.

\bibitem[\protect\citeauthoryear{Xue \bgroup \em et al.\egroup
  }{2022}]{xue2022byt5}
Linting Xue, Aditya Barua, Noah Constant, Rami Al-Rfou, Sharan Narang, Mihir
  Kale, Adam Roberts, and Colin Raffel.
\newblock {Byt5: Towards a token-free future with pre-trained byte-to-byte
  models}.
\newblock {\em Transactions of the Association for Computational Linguistics},
  10:291--306, 2022.

\bibitem[\protect\citeauthoryear{Zhang \bgroup \em et al.\egroup
  }{2014}]{zhang2014ontology}
Yuanzhe Zhang, Xuepeng Wang, Siwei Lai, Shizhu He, Kang Liu, Jun Zhao, and
  Xueqiang Lv.
\newblock {Ontology matching with word embeddings}.
\newblock In {\em Chinese computational linguistics and natural language
  processing based on naturally annotated big data}, pages 34--45. Springer,
  2014.

\end{thebibliography}

\end{document}